\documentclass[fleqn,10pt,twocolumn]{AROB-ISBC-SWARM24}

\title{Development of a Practical Articulated Wheeled In-pipe Robot for Both 3-4 in Force Main Inspection of Sewer Pipes}

\author{Kenya Murata${}^{1\dagger}$ and Atsushi Kakogawa${}^{2}$}
\speaker{Kenya Murata}

\affils{${}^{12}$Department of Robotics, Ritsumeikan
University, Shiga, Japan\\
(E-mail: ${}^{1}$rr0112rk@ed.ritsumei.ac.jp, ${}^{2}$kakogawa@fc.ritsumei.ac.jp)\\
}

\abstract{%
This paper reports a practical articulated wheeled in-pipe inspection robot ``AIRo-7.1'' which is waterproof and dustproof, and can adapt to 3 to 4 in inner diameters. The joint torque can be adjusted by a PWM open-loop control. The middle joint angle can be controlled by a position feedback control system while the other two joints are bent by torsional springs. Thanks to this simple and high-density design, not only downsizing of the robot but also wide range of the adaptive inner diameter were achieved. However, the relationship between the actual middle joint torque value and the PWM duty ratio should be pre-known because the reducer used in AIRo-7.1 was designed by ourselves. Therefore, preliminary experiments were conducted to clarify the relationship between them. To examine the adaptive movement, experiments in both 3 in and 4 in pipes with vertical, bend, and diameter change sections. Finally, field experiment was also conducted. From the results, high adaptability to different inner diameters of pipes and slippery environments were confirmed although waterproof and dustproof were not perfectly working.}

\begin{document}

\maketitle

\section{INTRODUCTION}\label{intro}
In recent years, the aging of water supply facilities has become more serious, and burst and leakage accidents have begun to occur. In paticular, a sewer pipe break leads to gradually erode the soil and may eventually cause a subsidence accident. The risk is even more serious because sewage contains hydrogen sulfide which increases the health hazards. As of 2021, 480,000 km of sewage pipes have already been laid in Japan, and 13,000 km of them are more than 50 years old. This number is expected to increase rapidly in the near future.

In response to this situation, the Ministry of Land, Infrastructure, Transport and Tourism in Japan established maintenance and repair standards based on the sewerage law in 2015. It made mandatory to inspect sewer pipes at least once every five years. Some inspection robots have been already commercialized such as scope cameras and self-propelled inspection equipments \cite{viet} \cite{jana}. However, these products are not available for all sewer facilities.

\begin{figure}[tbp]
\centering
\includegraphics[width=80mm]{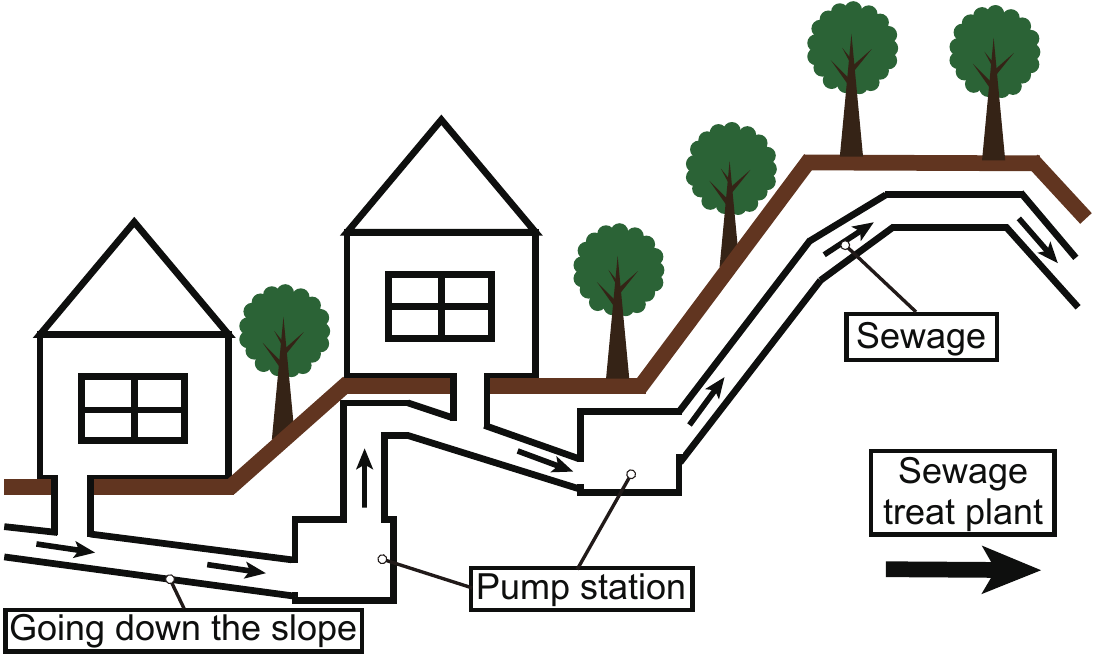}
\caption{Image of ``force main''.}
\label{force_main}
\end{figure}

For example, there are special pipes called ``force main'' (Fig. \ref{force_main}) that pumps up the sewage to a higher level. Force main is used in the places in which the wastewater cannot be carried naturally by gravity. It is reported that the total length of the force main in Japan is approximately 10,000 km. To increase water pressure, the inner diameters of the general force main are designed to small; 2 in, 3 in, and 4 in (approximately 50 mm, 75 mm, and 100 mm, respectively). Note that the diameter of some force mains exceed 4 in because of the regional differences \cite{new}. In addition, there are many bends and vertical sections in a short range. Therefore, it has been impossible to inspect them using a conventional products due to the size limitation and maneuverability limitation to three-dimensionally winding pathway.

On the other hand, there have been a few reports of inspection robots for the force main so far. Kubota Corporation (2-47, Shikitsuhigashi 1-chome, Naniwa-ku, Osaka Japan) has developed a push-in inspection device called ``Snei-kun" \cite{sneikun}. Kantool Corporation (1-3, Yosiyanipponbasi Bld.8th floor, Nihombashikodemma-cho, Chuo-ku, Tokyo, 103-0001, Japan) has commercialized a push-in camera specifically designed for small-diameter sewer pipes called ``Agilios'' \cite{kantool}. Nakamura et al. and Solaris Corporation (14-13, Higashiyama-cho, Itabashi-ku, Tokyo, 174-0073, Japan) have also proposed a pneumatically driven inspection robots with peristaltic locotion\cite{nakamura}. Koei Dream Works (Zizouyamasita 2068, Kazama, Yamagata, 990-2221, Japan) provides inspection services using a water hydraulic driven inspection device called ``Haikan Kun Type II" \cite{koei}.

However, the push-in type device cannot proceed to the depths of the winding pipes by itself due to the cable friction as mentioned above. Peristaltic locomotion has the disadvantage of slow moving speed of 0.03 m/s and less \cite {sola}. Pneumatic and hydraulic actuators may achieve drastic downsizing of the robot. However, they require air compressor and high-water-pressure equipment, thus, not easy to use in the field quickly. 

On the other hand, high adaptability pipe inspection robots with a structure combining multi-link and active wheel drive have been reported \cite{ot}-\cite{pipetron}. Although the inner diameters are still limited to 3–8 in, we think that this type is the fastest and the most adaptable to a variety of pipe shapes in size limitation. Our previous researches have also focused on this and showed some maneuverability with real equipments \cite{adv}-\cite{tie}.

\begin{figure}[tbp]
\centering
\includegraphics[width=80mm]{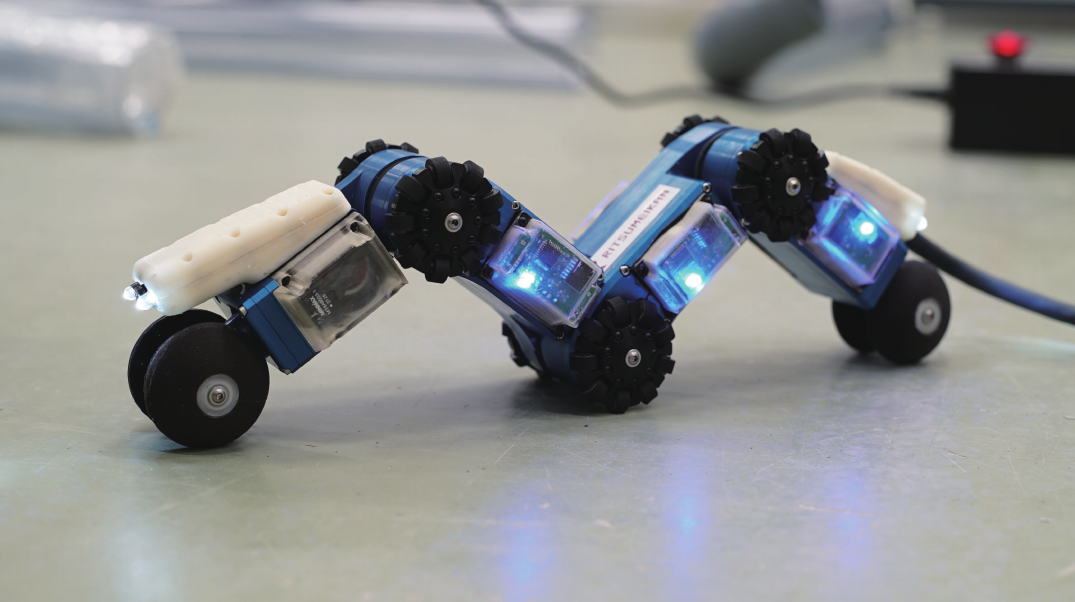}
\caption{Overview of an in-pipe inspection robot with waterproof and dustproof: AIRo-7.1.}
\label{airo7.1}
\end{figure}

This paper proposes our latest in-pipe robot, AIRo-7.1 (Fig. \ref{airo7.1}), which is driven by geared electric motors with improvements of waterproof and dustproof for traveling in force mains. This robot can pass through straight and bend pipes with both 3 in and 4 in inner diameter and even adapt to joint part of them such as increaser/decreaser. This adaptive movement is achieved by a simple open-loop joint torque adjustment (PWM control). However, the relationship between the actual torque value and the PWM duty ratio should be pre-known. 
\section{DESIGN OF THE IN-PIPE INSPECTION ROBOT}\label{mech}
\subsection{Mechanical design}
\begin{figure}[tbp]
\centering
\includegraphics[width=85mm]{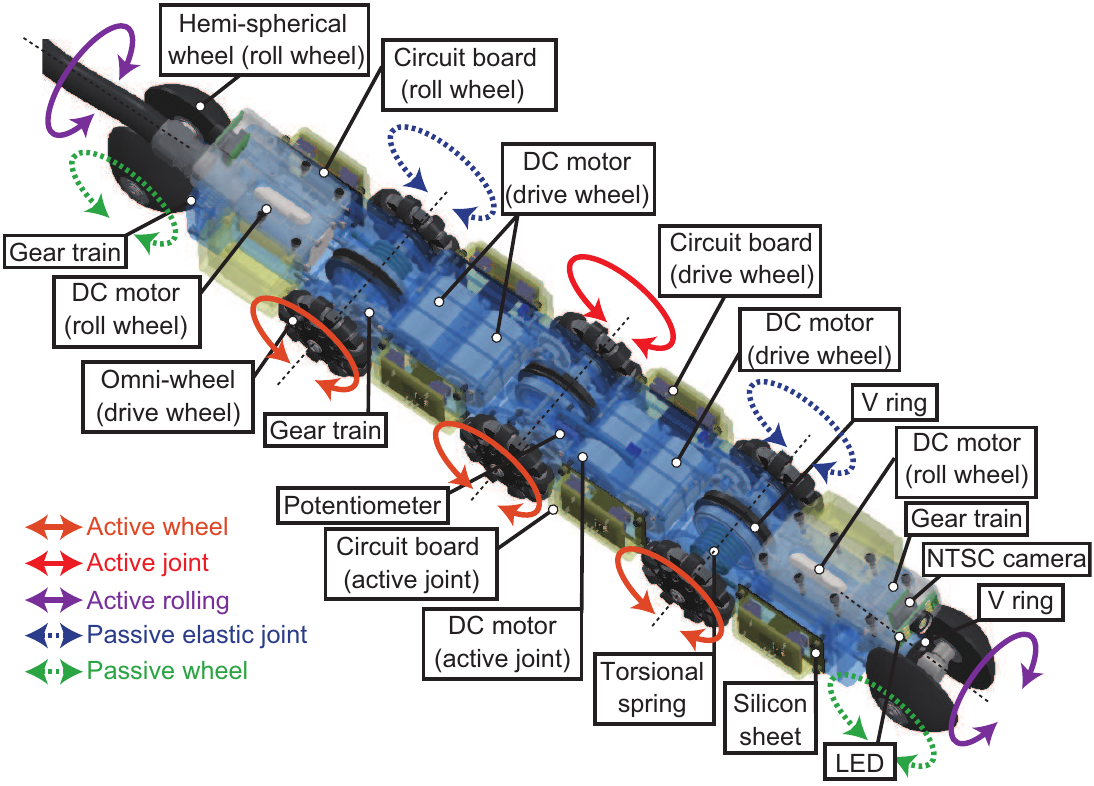}
\caption{CAD model of the in-pipe inspection robot AIRo-7.1.}
\label{7.1straight}
\end{figure}

The CAD model of AIRo-7.1 is shown in Fig. \ref{7.1straight} and the specifications are given in Table \ref{spec}. 
Specifications for moving speed, traction force, and joint torque are estimates calculated from the motor and gears.
AIRo-7.1 is based on our previous robot AIRo-5.2 \cite{tie}. It has three-joint, four-link structure with omni-wheels for forward and backward movement at each joint and spherical wheels for rolling rotation at the front and rear ends \cite{tadakuma}. As in the AIRo-5.2, two torsion coil springs are installed in the two front and rear joints, while the middle joint rotates actively with a geared motor. NTSC cameras of approximately 1.2 megapixels are mounted at the front and rear ends, allowing the user to see the in-pipe images.

\begin{table}[tbp]
	\caption{Specification of AIRo-7.1}
	\label{spec}
		\centering
			\begin{tabular}{|c|c|}
                        \hline
			Adaptive pipe inner diameter & 3-4 in\\
                        \hline
			Total length (when extended) & 0.51 m\\
                        \hline
			Total weight (without cable) & 1.57 kg\\
			\hline
			Max. moving speed & 0.088 m/s\\
			\hline
			Max. continuous traction force & 151 N (15.4 kgf)\\
			\hline
			Peak traction force & 728 N (75.2 kgf)\\
			\hline
			Max. continuous joint torque & 2.56 Nm\\
			\hline
			Peak joint torque & 12.32 Nm\\
			\hline
			Nominal voltage & DC 24 V\\
			\hline
			Communication &\begin{tabular}{c} Controller \\Area Network (CAN)\end{tabular}\\
			\hline
			\end{tabular}
	\end{table}
        
Although the minimum adaptable inner diameter has been reduced from 4 in to 3 in, the robot can still pass through a 4 in inner diameter. Since many of the existing small in-pipe inspection robots have been designed to pass through only certain inner diameters, the adaptability to inner diameter change of 1 in (approximately 25\% of 4 in) can be considered to be superior for practical use.

\begin{figure}[tbp]
\centering
\includegraphics[width=90mm]{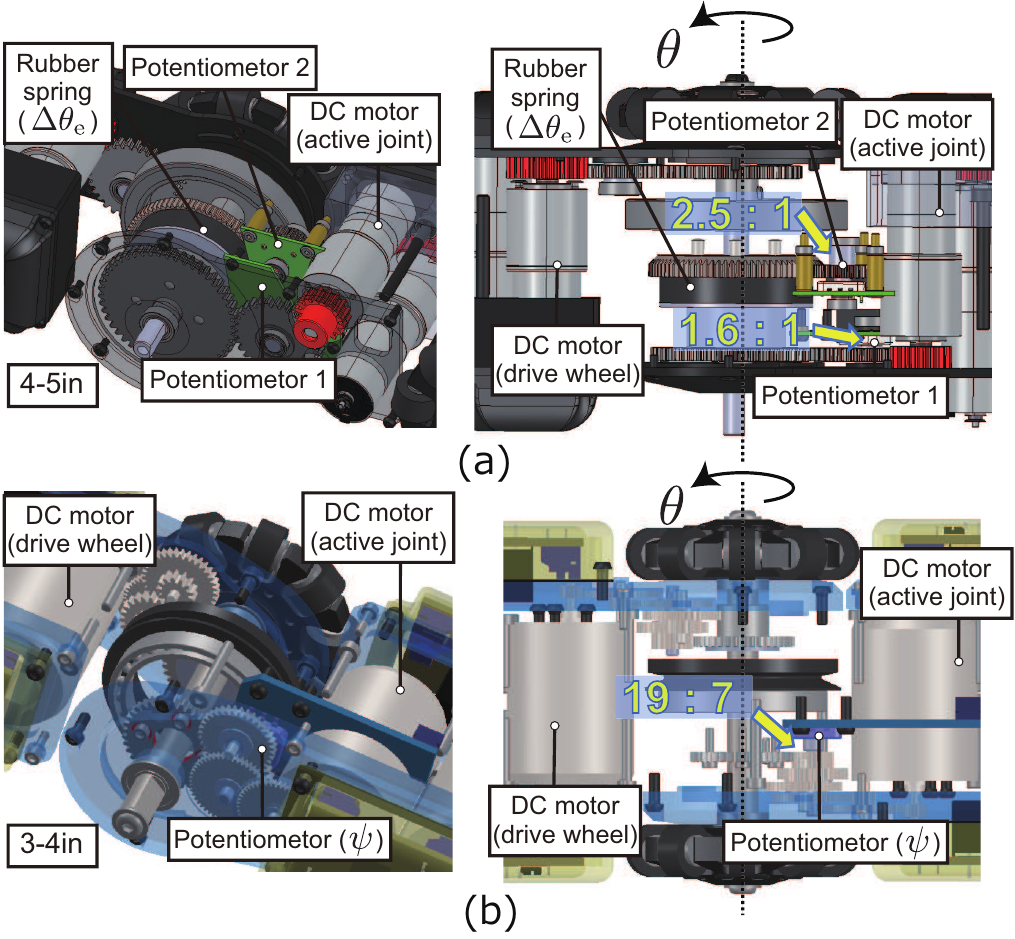}
\caption{Modification of active middle joints due to miniaturization (a) AIRo-5.2 (4-5 in) (b) AIRo-7.1(3-4 in).}
\label{middlejoint}
\end{figure}

The structure of the middle active joint of AIRo-5.2 (the older model with an adaptive inner diameter of 4-5 in) and AIRo-7.1 (the newer model with an adaptive inner diameter of 3-4 in) is shown in Fig. \ref{middlejoint}. In AIRo-5.2, a SEA (series elastic actuator) \cite{sea} was installed in the middle joint, and thus, both angle and torque can be controlled.

\begin{figure}[tbp]
\centering
\includegraphics[width=80mm]{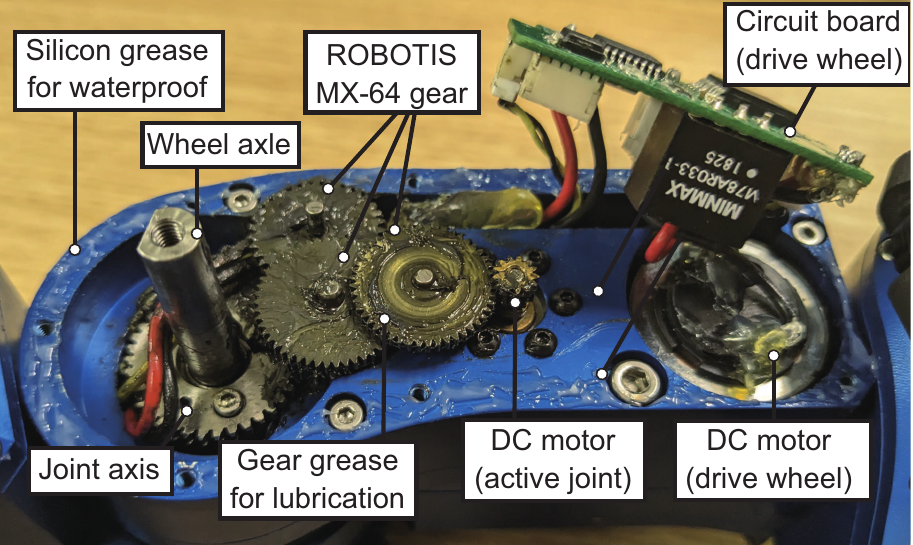}
\caption{Transmission mechanism of active middle joint.}
\label{gear}
\end{figure}

In AIRo-7.1, the joint torque is controlled by open-roop PWM control without using any SEA and torque sensor. As in the AIRo-5.2, the joint angle is controlled by a position feedback control system using a potentiometer. The transmission mechanism has also been modified, consisting only a simple and tiny spur gear train with reduction ratio of approximately 200 (MX-64 provided by ROBOTIS CO., LTD, 37, Magokjungang 5-ro 1-Gil, Gangseo-gu, Seoul, Korea), and the gear grease for lubrication reduces the friction loss (Fig. \ref{gear}). The uniform motor and reduction system are installed in the omni-wheels and the spherical wheels. For waterproofing and dustproofing, silicone grease or silicone sheets were applied and installed to the gaps between the parts.

\subsection{Communication system}
\begin{figure}[tbp]
\centering
\includegraphics[width=85mm]{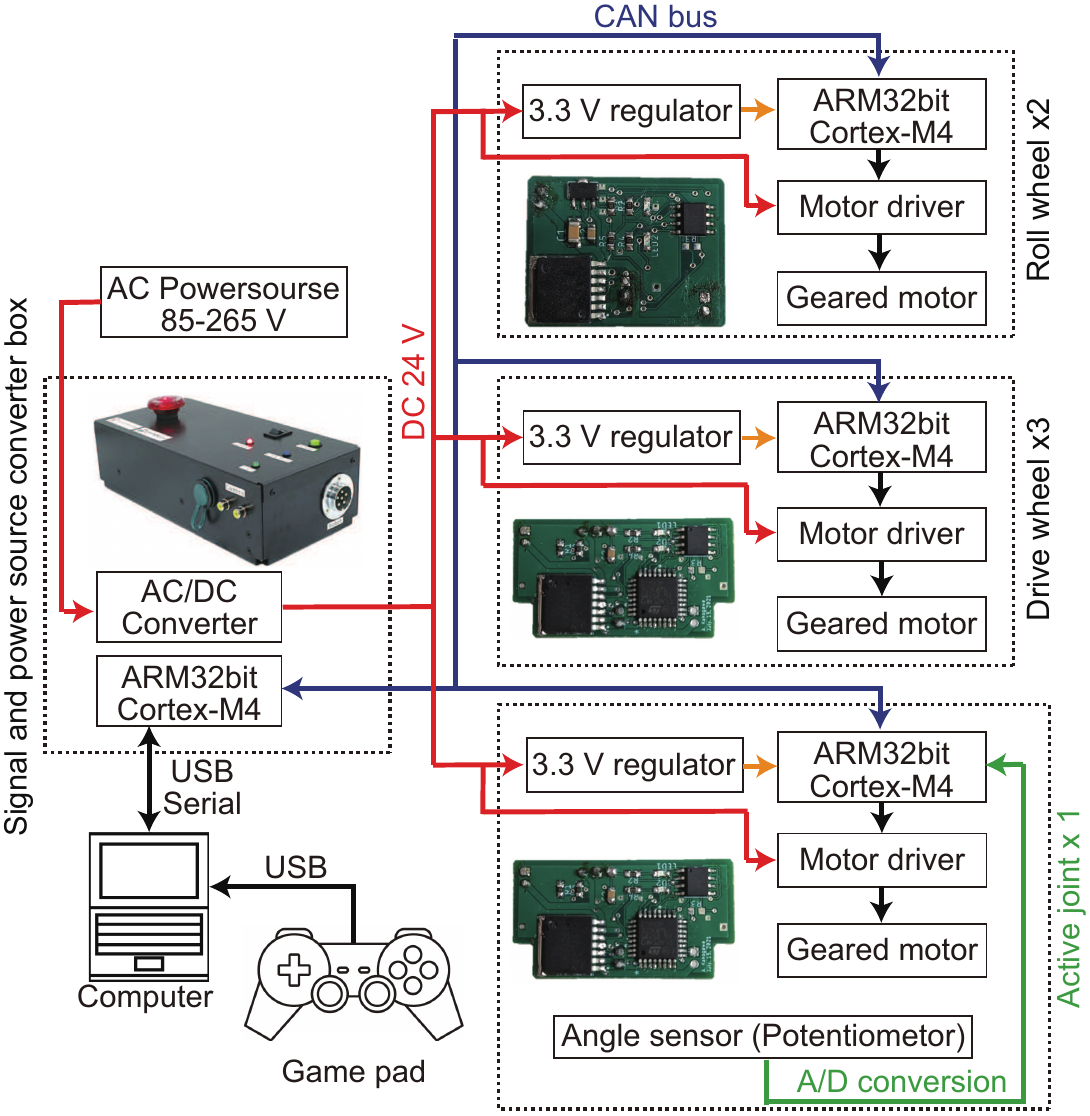}
\caption{Communication system of AIRo-7.1.}
\label{can}
\end{figure}

The communication system used for the AIRo-7.1 is shown in Fig. \ref{can}. Two types of our designed printed circuit board are used for the drive wheels, the active joint, and the roll wheels. The same circuit board is used for the drive wheels and the active joint. The ARM 32-bit microcontroller (Cortex-M4) is used for the motor control and communication. The CAN-BUS protocol (Controller Area Network) was adopted to send a command and to receive a signal from the robot. The target signals (moving forward and backward, rolling around the pipe axis, stop and target angle) are sent through a computer and a signal and power source converter box (from the USB serial to CAN-BUS). This box has an AC/DC converter of the power source and an EMO (Emergency Off) button. 
By adjusting the PWM duty ratio, the speed and middle joint torque of the drive and roll wheels can be changed. The BD6232HFP-TR (ROHM Co. Ltd.,21 Saiin Mizosaki-cho, Ukyo-ku, Kyoto 615-8585 Japan) is also used to change the direction of the motor rotation and to amplify the motor voltage. The maximum output current of each circuit board is 2.0 A despite its small size. The microcontroller reads the signals from the potentiometer through a voltage divider (from 5 V to 3.3 V). This is because the nominal voltage of the potentiometer RDC506018A is 5 V but the VDD of the microcontroller (IC power-supply voltage) is 3.3 V. 

\section{OPEN-LOOP JOINT TORQUE CONTROL}\label{control}
\begin{figure}[tbp]
\centering
\includegraphics[width=80mm]{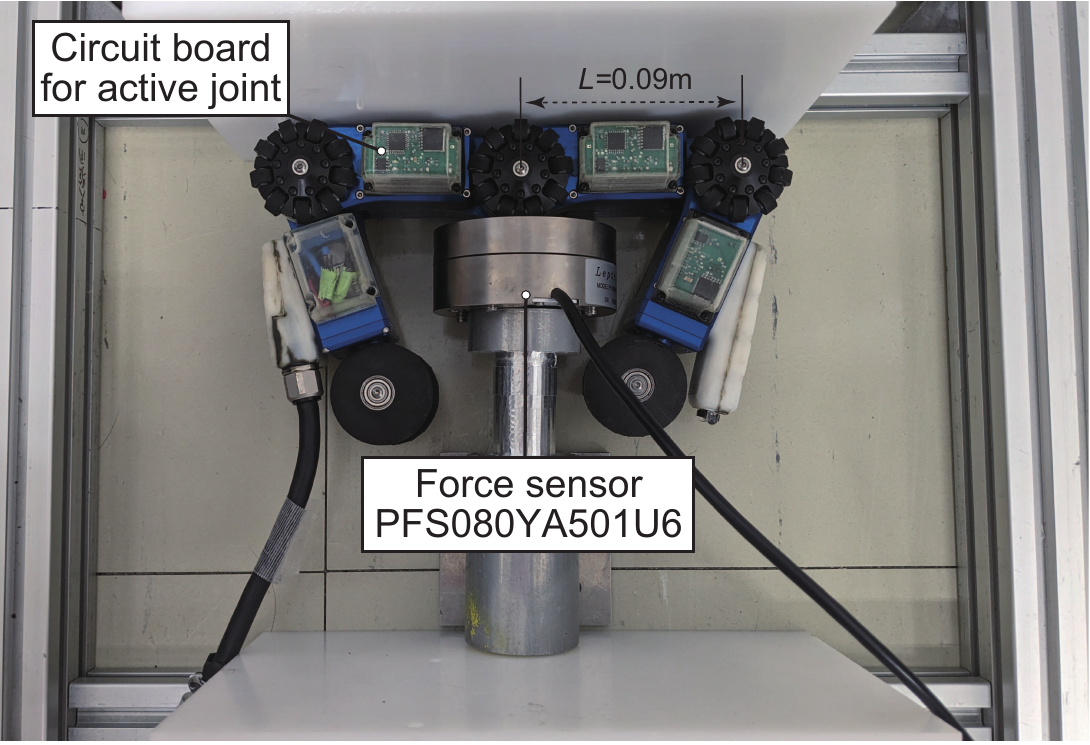}
\caption{Experimental setup.}
\label{testpwm}
\end{figure}

Use the force sensor PFS080YA501U6 (Leptrino Co., Ltd. Saku, Nagano, Japan) to verify the relationship between the actual middle joint torque value and the PWM duty ratio.
Fig. \ref{testpwm} shows the test rig. The measured joint torque ($\tau_{\mathrm{sens}}$) was calculated by the static equilibrium, the link length between the drive wheels ($L$) and the measured force ($F_{\mathrm{sens}}$) by $\tau_{\mathrm{sens}}$ = $F_{\mathrm{sens}}L/2$. 
The duty ratio increased from 0 \% to 100 \% and then decreasing from 100 \% to 0 \%. The sampling time was 50 ms, the input varied by 0.2 \% and was measured twice.

\begin{figure}[tbp]
\centering
\includegraphics[width=85mm]{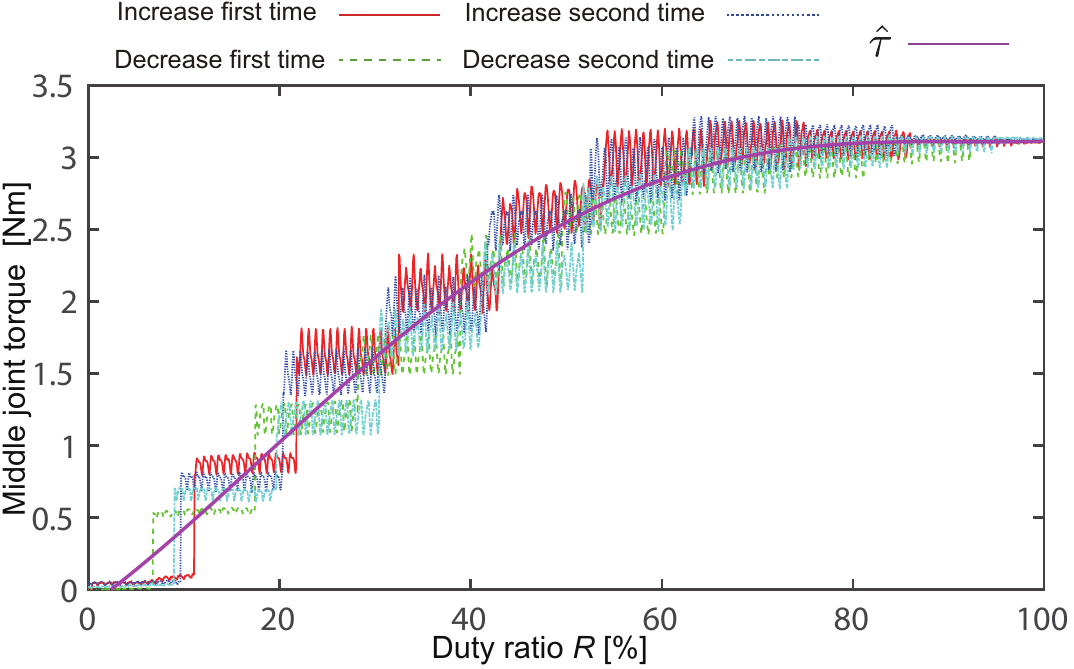}
\caption{Joint torque measured by the force sensor ($\hat{\tau}$: simulated middle joint torque).}
\label{Nm_pwm}
\end{figure}

Fig. \ref{Nm_pwm} plots the torque measurement results.
It can be seen that the measured torque increases as the duty ratio increases, but the torque does not change around 3 Nm after 70 \%. This indicates that the torque control by PWM control is not linear.
In addition, the although open-roop input is a linear fashion, the measured joint torque $\tau_{\mathrm{sens}}$ varies in a stepwise fashion. This is expected because the gear is slightly elastically deformed in a stepwise fashion, which is measured as a change in torque.

Using the results of Fig. \ref{Nm_pwm}, the estimated torque ($\hat{\tau}$) is as follows:
\begin{equation}
\hat{\tau}=AR^4+BR^3+CR^2+DR+E
\label{tau}
\end{equation}
	
$A (=-0.1178)$, $B (=4.7894\times10^{-2})$, $C (=7.6041\times10^{-4})$, $D (=-1.6902\times10^{-5})$, $E (=-7.7385\times10^{-8})$ are all constants and were fitted by the least squares method and $R$ is duty ratio. The joint torque with respect to duty ratio can be estimated from this equation (\ref{tau}). 
The estimated max. continuous joint torque of this robot is 2.56 Nm, so a duty ratio of 50\% or less is used.

\section{EXPERIMENTS}\label{experiments}
In this section, traveling experiments are used to verify that open-loop joint torque control can be adapted to various types of pipes. In the following experiments, the joint torque is calculated by equation (\ref{tau}).
\subsection{Preliminary experiment}
\begin{figure}[tbp]
\centering
\includegraphics[width=80mm]{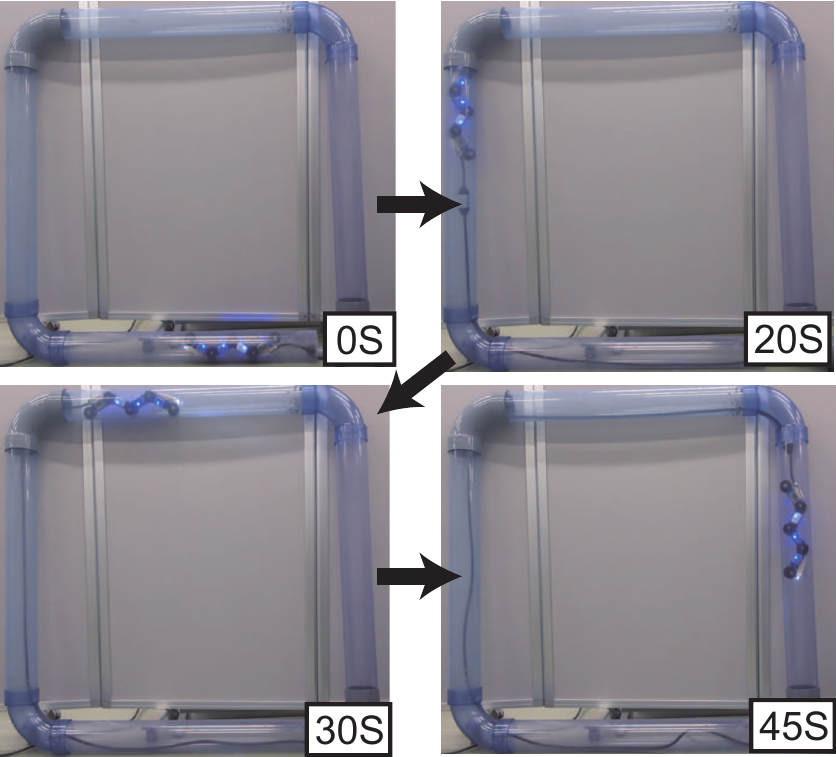}
\caption{Vertical 3 in pipe travel.}
\label{pre3in}
\end{figure}

\begin{figure}[tbp]
\centering
\includegraphics[width=80mm]{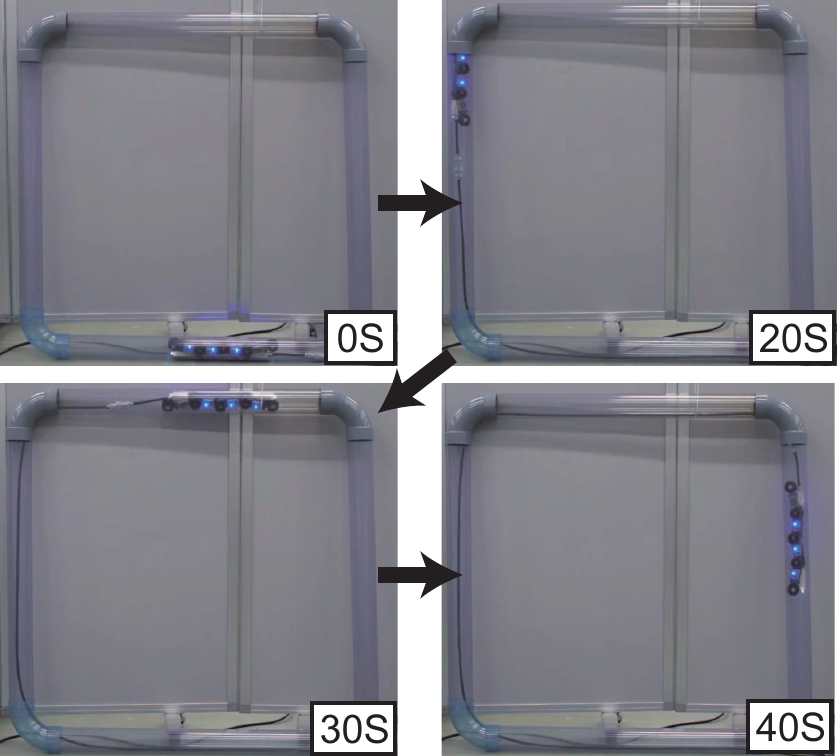}
\caption{Vertical 4 in pipe travel.}
\label{pre4in}
\end{figure}

First, two types of pipe courses were prepared: a 1 m 3 in polyvinyl chloride (PVC) pipe and four 4 in pipes, each connected by three 90 degree bent pipes (Fig \ref{pre3in}, Fig \ref{pre4in}). 
Pipes with a radius of curvature of 0.1 m and 0.128 m were used for the 3 in and 4 in 90 degree bent pipes, respectively.
Initially, this robot was driven with a joint torque of approximately 0.42 Nm (10 \% duty ratio), but the drive wheels slipped on the bent pipe. Therefore, a traveling test was performed with a joint torque of approximately 1.32 Nm (25 \% duty ratio). 
It was confirmed that AIRo-7.1 can easily pass through both 3 in and 4 in pipes  including bent and vertical pipes.

\begin{figure}[tbp]
\centering
\includegraphics[width=80mm]{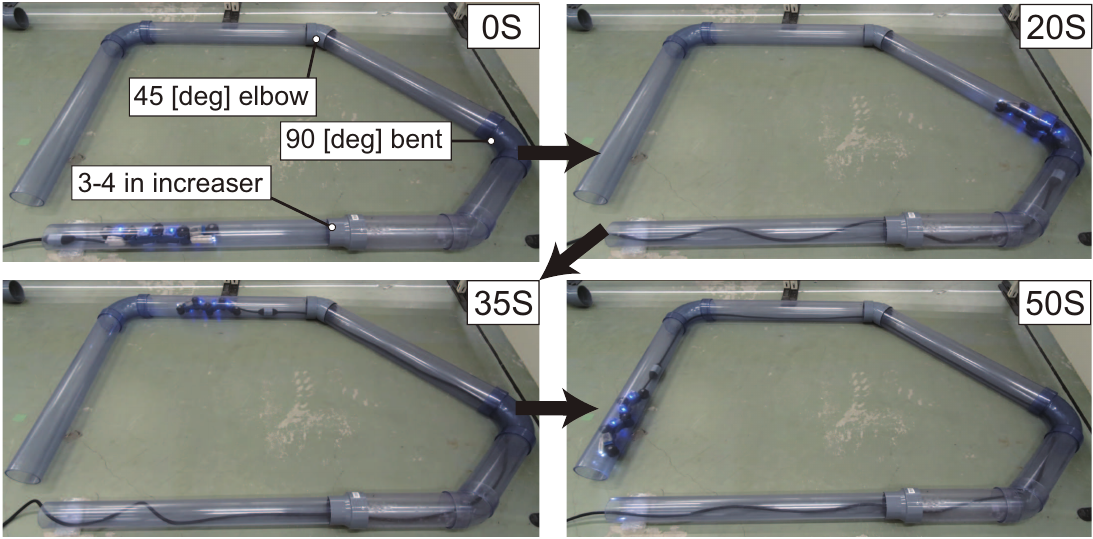}
\caption{3-4 in increaser, 90 degree bent and 45 degree elbow pipe travel.}
\label{increaser}
\end{figure}

\begin{figure}[tbp]
\centering
\includegraphics[width=80mm]{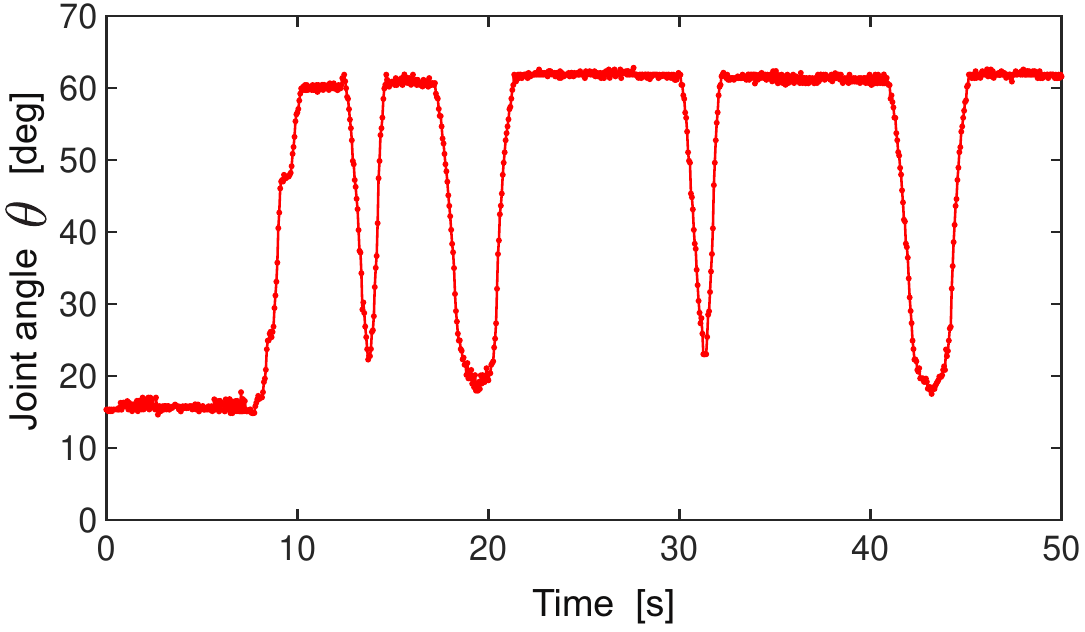}
\caption{Measured middle joint angle $\theta$.}
\label{increaserdata}
\end{figure}

Next, a traveling experiment was performed on a 3-4 in pipe containing an ``increaser pipe'' with varying pipe diameters. Fig. \ref{increaser} shows the experimental environment, and Fig. \ref{increaserdata} plots the angle read by the potentiometer. The 45 degree elbow pipe with a radius of curvature of 0.062 m is used. This experiment was also performed with the joint torque of approximately 1.32 Nm (duty ratio 25 \%). The results of the traveling experiments show that the robot successfully passed through the increaser pipe, 90 degree bent pipes and 45 degree elbow pipes. 

\subsection{Field experiment}
To verify the performance of AIRo-7.1 in actual environments for practical applications, a field test was conducted in a force main with a nominal inner diameter of 75 mm. Since this was the primary field test, the traveling distance for the verification was only 3 m. 

\begin{figure}[tbp]
\centering
\includegraphics[width=80mm]{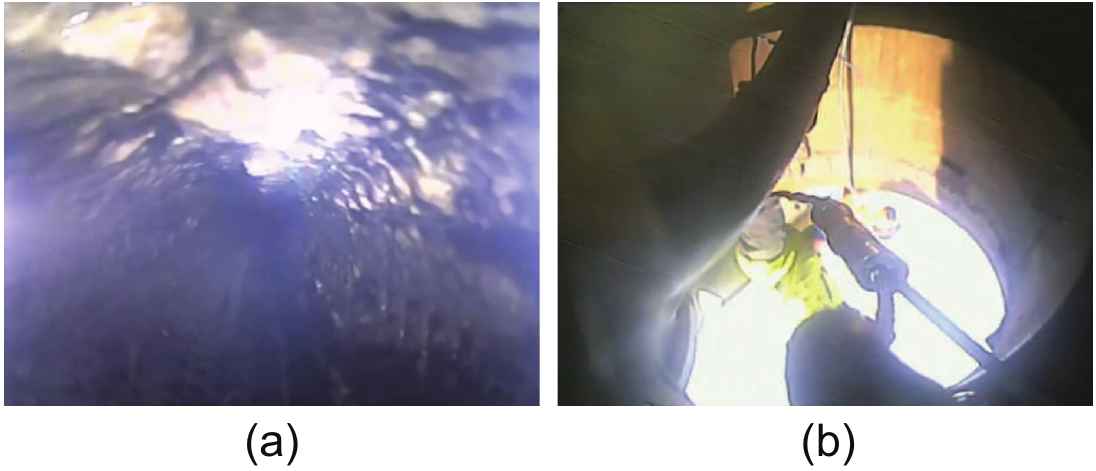}
\caption{(a) Front camera images (b) Rear camera images.}
\label{exp_camera}
\end{figure}

Initially, the joint torque was 1.32 Nm (25 \% duty ratio). However, as shown in Fig. \ref{exp_camera}, the inner wall of the force main was covered with a large amount of sewage, which caused the drive wheels to slip. Therefore, by increasing the joint torque to 2.55 Nm (50 \% duty ratio), the drive wheels travel without slipping. This result shows that the joint torque control can be adapted to slippery environments. However, as shown in Fig. \ref{exp_resin} the circuit board attached to the side of AIRo-7.1 generated heat and softened the resin case due to the increased current to increase the torque and continuous control for more than 15 minutes. As a result, sewage flowed in through the gaps.

\begin{figure}[tbp]
\centering
\includegraphics[width=70mm]{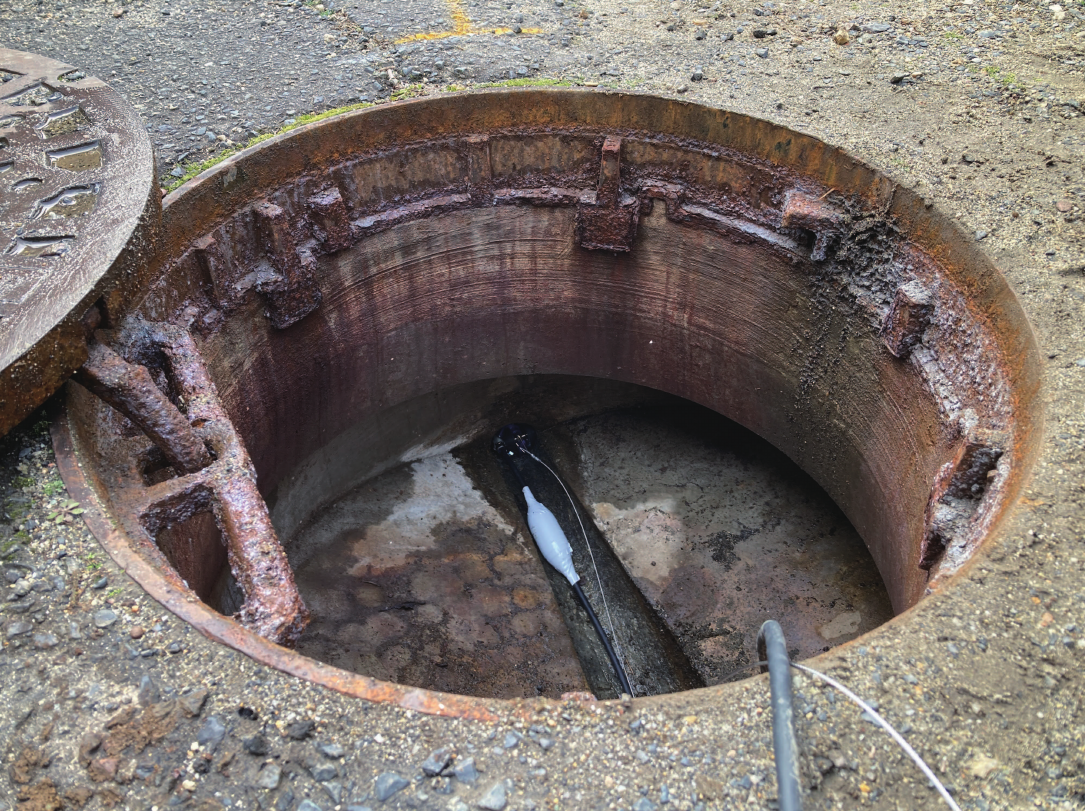}
\caption{AIRo-7.1 entering a force main in sewer pipes.}
\label{exp_inpipe}
\end{figure}

\begin{figure}[tbp]
\centering
\includegraphics[width=70mm]{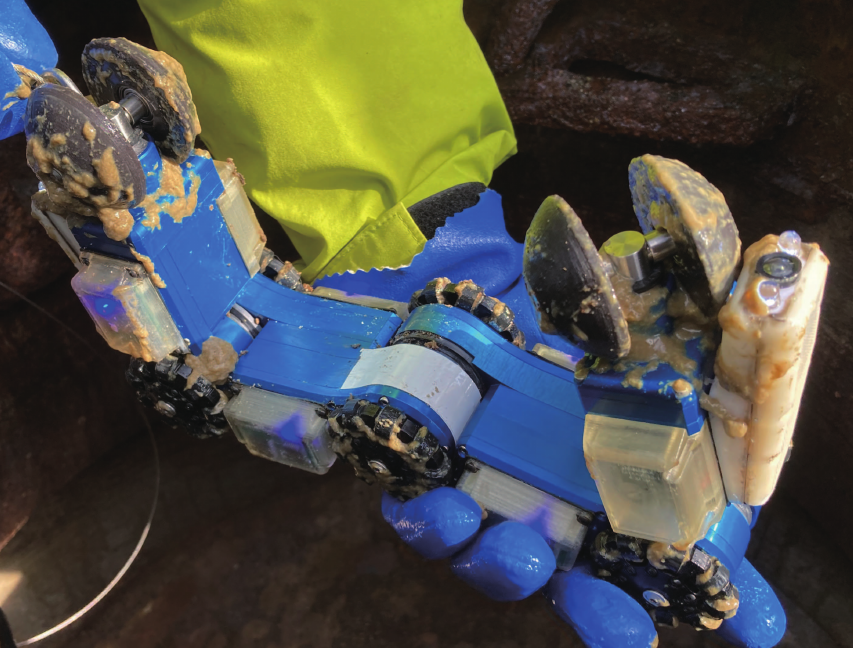}
\caption{AIRo-7.1 after the experiment.}
\label{exp_robot}
\end{figure}

\begin{figure}[tbp]
\centering
\includegraphics[width=70mm]{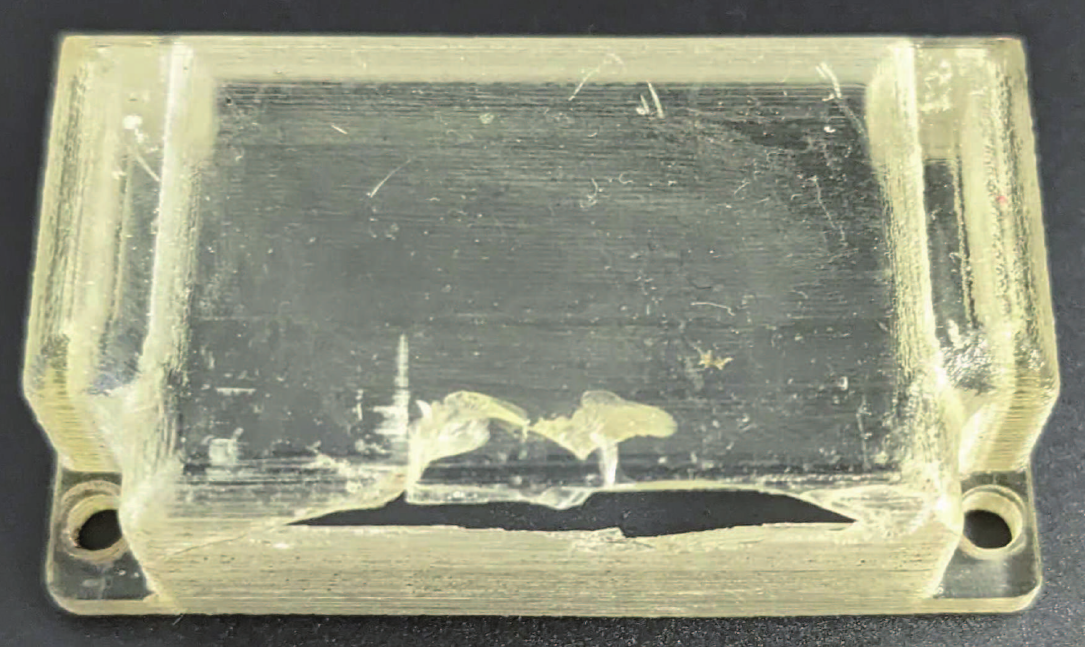}
\caption{Softened and broken resin case.}
\label{exp_resin}
\end{figure}

\begin{figure}[tbp]
\centering
\includegraphics[width=70mm]{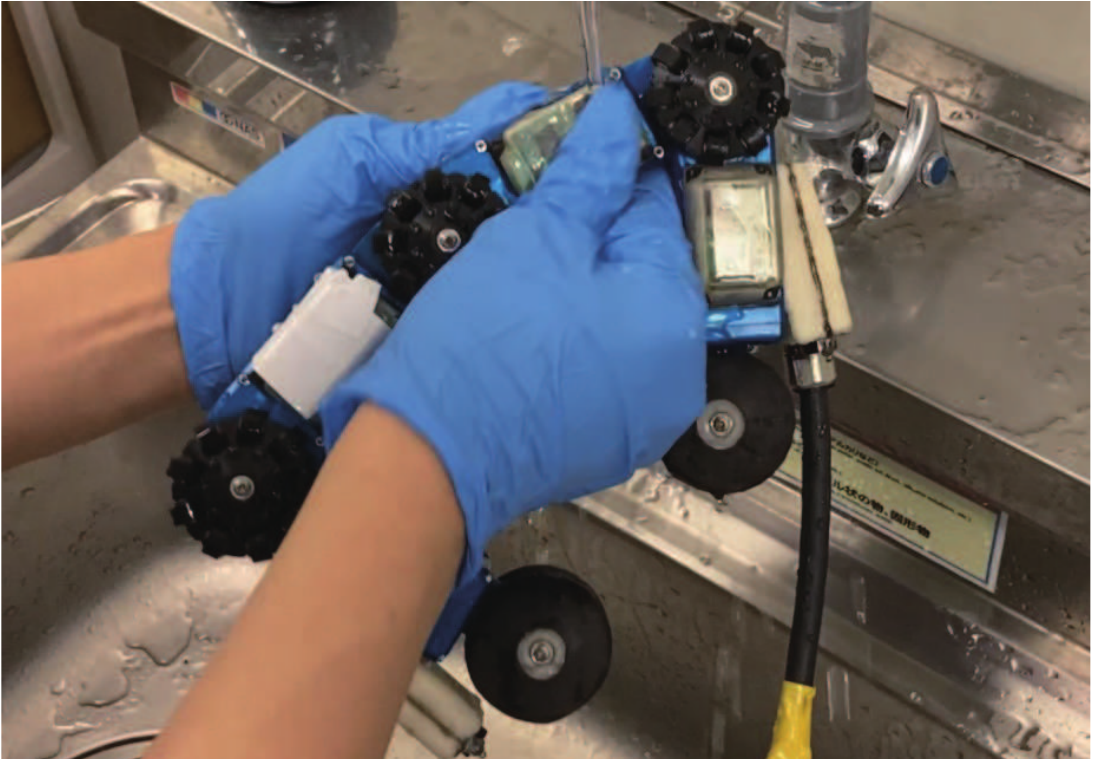}
\caption{Washed AIRo-7.1.}
\label{exp_resin}
\end{figure}

\section{CONCLUSIONS}\label{con}
This study proposed the development of a practical articulated wheeled in-pipe robot for 3-4 in force main inspection in sewer pipes. First, in order to perform the sewer force main inspection, which was the goal of this project, it was necessary to travel with sewage remaining and be able to travel 3 in pipes. Therefore, a new design was created with a waterproof and dustproof design and a 1 in reduction in size (approximately 25\% of 4 in). For waterproofing and dustproofing, silicone grease or silicone sheets were applied and installed to the gaps between the parts. On the other hand, the former design active middle joint structures are difficult to miniaturize. Therefore, AIRo-7.1 was changed to a PWM open-loop joint torque control, using only geared motors and potentiometers instead of any SEA and torque sensor. However, the relationship between the actual torque value and the PWM duty ratio should be pre-known to prevent the gear from breakage due to overload. Therefore, preliminary experiments were conducted to derive approximate the joint torque with the duty ratio.

In the preliminary experiment, 3 in and 4 in straight pipes and bent pipes were connected and set up vertically to check the passage performance. In addition, a traveling experiment was also conducted using an increaser pipe with a pipe diameter varying from 3 to 4 in. As a result, the robot passed smoothly through each pipes, showing the effectiveness of joint torque control. In addition, this is very important for practical use because the robot can adapt even to a change in the inside diameter of approximately 1 in (approximately 25\% of 4 in), which has been difficult for most small in-pipe inspection robots.

Field experiments showed that the system can be adapted to slippery environments by increasing the joint torque. However, the circuit board was damaged due to heat generation, and we plan to take measures against heat generation and improve the cover. In addition, a 1.2 megapixel NTSC camera was installed in this experiment, but it could not clearly record the situation inside the sewer pipes. Therefore, a USB camera with higher resolution will be used in the next experiment.

\addtolength{\textheight}{0cm}   



\section*{ACKNOWLEDGMENT}
We would like to express our sincere thanks to Mr. Takayasu Yamamoto, Kusatsu City Hall, and Mr. Tsuyoshi Hirano, Daigo Sangyo  Corporation. 

This work was supported by *Shiga Prefectural Subsidy for the Promotion of Social Implementation of Near-Future Technologies, etc and *Ritsumeikan Impact-Makers Inter X (Cross) Platform Commercialization Grants.




\begin{thebibliography}{99}
\bibitem{viet}
T. Nguyen, P. Nguyen, and T. Tuong, 
``A study of pipe-cleaning and inspection robot'',
IEEE International Conf. Robotics and
Biomimetics, pp. 2593-2598, 2011.

\bibitem{jana}
J. Knedlová, O. Bilek, D. Sámek, and P. Chalupa, 
``Design and construction of an inspection robot for the sewage pipes'', in Proc. The Multi-Conf. Engineering and Technology Innovation, vol. 121, 2017.

\bibitem{new}
2018 Annual report of the institute for new technology in sewerage, pp. 1-12, 2018.

\bibitem{sneikun}
Kubota Corporation, Annual Securities Report for the 129th business term (From January 1, 2018 to December 31, 2018), p.21.

\bibitem{kantool}
Kantool Corporation, https://kantool.co.jp (access: 11, 2023)

\bibitem{nakamura}
T. Kishi, M. Ikeuchi, and T. Nakamura, ``In-pipe inspection robot capable
of actively exerting propulsive and tractive forces with linear antagonistic
mechanism'', IEEE Access, vol. 9, pp. 131245–131259, 2021.

\bibitem{koei}
Koei Dream Works, https://koeidreamworks.jp/service/robot-two (access: 11, 2023)

\bibitem{sola}
Solaris inc, https://solaris-inc.com/products/sooha (access: 11, 2023)

\bibitem{ot}
J. Borenstein, M. Hansen, and A. Borrell, ``The OmniTread OT-4 serpentine
robot-design and performance'', J. Field Robot., vol. 24, no. 7,
pp. 601–621, 2007.

\bibitem{vis}
H. Schempf, et al., ``Visual and nondestructive evaluation inspection of live
gas mains using the ExplorerTM family of pipe robots'', J. Field Robot.,
vol. 27, no. 3, pp. 217–249, 2010.

\bibitem{jang}
H. Jang et al., ``Development of modularized in-pipe inspection robotic
system:MRINSPECT VII, Robotica'', vol. 40, no. 5, pp. 1361–1384, 2021.

\bibitem{foster}
G. C. Vradis and W. Leary, ``Development of an inspection platform and a
suite of sensors for assessing corrosion and mechanical damage on unpiggable
transmission mains'', Technical Report of NGA and Foster-Miller, 2004.

\bibitem{gas}
E. Dertien, S. Stramigioli, and K. Pulles, ``Development of an inspection
robot for small diameter gas distribution mains'', in Proc. IEEE Int. Conf.
Robot. Automat., 2011, pp. 5044–5049.

\bibitem{pipetron}
P.~Debenest, M.~Guarnieri, and S.~Hirose,
``PipeTron Series - Robots for Pipe Inspection'',
in Proc. the 3rd Int. Conf. Applied Robotics for the Power Industry, pp. 1-6, 2014.

\bibitem{adv}
A. Kakogawa and S. Ma, ``Design of a multilink-articulated wheeled
pipeline inspection robot using only passive elastic joints'', Adv. Robot.,
vol. 32, no. 1, pp. 37–50, 2018.

\bibitem{tie}
A. Kakogawa, K. Murata, and S. Ma, ``Automatic T-Branch Travel of an Articulated Wheeled In-Pipe Inspection Robot Using Joint Angle Response to Environmental Changes'', IEEE Transactions on Industrial Electronics, vol.70, iss.7, pp.7041-7050, 2023.

\bibitem{tadakuma}
K. Tadakuma, ``Tetrahedral Mobile Robot with Novel Ball Shape Wheel'', in Proc. the First IEEE/RAS-EMBS Int. Conf. Biomedical Robotics and Biomechatronics, pp. 946-952, 2006.

\bibitem{sea}
G.A.~Pratt and M.~Williamson,
``Series Elastic Actuators'',
in Proc. the IEEE/RSJ Int. Conf. Intelligent Robots and Systems, pp. 399-406, 1995.


\end{thebibliography}
\end{document}